{

\documentclass[journal]{IEEEtran}

\usepackage[dvipsnames]{xcolor}
\usepackage{algorithm}
\usepackage{xspace}
\usepackage{algpseudocode}
\usepackage{multirow}
\usepackage{amsmath}
\usepackage{mathtools}
\usepackage{graphicx}
\usepackage{xcolor}
\usepackage{enumitem,kantlipsum}

\usepackage[normalem]{ulem}
\setlist{nolistsep}

\ifCLASSINFOpdf
\else
\fi
\hyphenation{op-tical net-works semi-conduc-tor}

\newcommand{\THIRDEA}{\textcolor{black}{SIEA-MCTS}}

\newcommand{\MCTSIts}{\textcolor{black}{Iterations}}

\newcommand{\SIEAMCTS}{\textcolor{black}{SIEA-MCTS}}

\newtheorem{mydef}{Def.}

 \usepackage{amssymb}
\usepackage{pifont}
\usepackage{tikz}

\usepackage{float,booktabs,array}

\begin{document}
%


\title{Towards Understanding the Effects of \\Evolving the MCTS UCT Selection Policy}

%
%
%

\author{Fred Valdez Ameneyro and Edgar Galv\'an$^*${\thanks{$^*$Leading and corresponding author.}} 
\thanks{Fred Valdez Ameneyro and Edgar Galv\'an are both with the Naturally Inspired Computation Research Group and with the Department
of Computer Science, Maynooth University, Lero, Ireland  e-mails:   fred.valdezameneyro.2019@mumail.ie and edgar.galvan@mu.ie}}

\markboth{}
\markboth{}

%



\maketitle

\begin{abstract}

  Monte Carlo Tree Search (MCTS) is a sampling best-first method to search for optimal decisions. The success of MCTS depends heavily on how the MCTS statistical tree is built and the selection policy plays a fundamental role in this. A particular selection policy that works particularly well, widely adopted in MCTS, is the Upper Confidence Bounds for Trees, referred to as UCT. Other more sophisticated bounds have been proposed by the community with the goal to improve MCTS’s performance on particular problems. Thus, it is evident that while the MCTS UCT behaves generally well, some variants might behave better. As a result of this, multiple works have been proposed to evolve a selection policy to be used in MCTS. Although all these works are inspiring, none of them have carried out an in-depth analysis shedding light under what circumstances an evolved alternative of MCTS UCT might be beneficial in MCTS due to focusing on a single type of problem. In sharp contrast to this, in this work we use five functions of different nature, going from a unimodal function, covering multimodal functions to deceptive functions. We demonstrate how the evolution of the MCTS UCT might be beneficial in multimodal and deceptive scenarios, whereas the MCTS UCT is robust in unimodal scenarios and competitive in the rest of the scenarios used in this study.

\end{abstract}

\begin{IEEEkeywords}
Monte Carlo Tree Search, UCT.
\end{IEEEkeywords}

%
\IEEEpeerreviewmaketitle

\section{Introduction}
 \label{sec:introduction}

Monte Carlo Tree Search (MCTS) is a sampling method for finding \textit{optimal decisions} by performing random samples in the decision space and building a tree according to partial results.  The evaluation function of MCTS relies directly on the simulation's outcomes.  The optimal search tree is guaranteed to be found with infinite memory and computation~\cite{kocsis2006bandit}. However, in more realistic scenarios, MCTS can produce good approximate solutions~\cite{Galvan_EnergyCon_2014,galvan2014heuristic}.

MCTS has gained popularity in two-player board games partly thanks to its success in the game of Go~\cite{alphago}, including beating professional human players. The diversification of MCTS in other research areas is extensive. For instance, MCTS has been explored in energy-based problems~\cite{Galvan_EnergyCon_2014,galvan2014heuristic} and in the design of deep neural network (DNN) architectures~\cite{wang2019alphax}. These two extreme examples, along with the applicability in games~\cite{9659930,DBLP:conf/ssci/AmeneyroGM20}, demonstrate the successful versatility, use and applicability of MCTS in different problem domains.

The success or failure of MCTS depends heavily on how the MCTS statistical tree is built. The selection policy, responsible for this, behaves  well when using the Upper Confidence Bounds for Trees (UCT)~\cite{10.1007/11871842_29}. Some conditions are to be met for this to work well, for example, the selection of a child of a given node is based on the exploration/exploitation trade-off. To this end, the UCT expression is normally used, yielding good results. Some of the parameters' values can be changed contributing to a better performing MCTS. Moreover, some sophisticated bounds have been proposed such as the single-player MCTS, adding a third term to the UCT formula and changing the value of a parameter~\cite{DBLP:journals/kbs/SchaddWTU12}. The UCB1-tuned also modifies the UCT expression to reduce the impact of the exploration term~\cite{DBLP:journals/ml/AuerCF02}. Thus, it is evident that while the UCT performs well on a range of problems, its adjustment or modification can have a more positive effect. 

It is clear then to see that the MCTS’s selection policy when adjusted properly can have a more positive impact on a given problem. Motivated by this, we use Evolutionary Algorithms (EAs)~\cite{EibenBook2003} to evolve the UCT formula. Specifically, we use a semantic-based approach to do so inspired by the positive impact that semantics has in Genetic Programming~\cite{Koza:1992:GPP:138936} as shown by multiple works~\cite{9872022,DBLP:conf/gecco/GalvanS19,DBLP:journals/gpem/UyHOML11} including our studies in Multi-Objective GP~\cite{Galvan:ASC:2021,DBLP:conf/ppsn/LopezMES16}.

{There are some interesting works using EAs in MCTS. Cazenave~\cite{Cazenave2007EvolvingMT} used GP to evolve the UCT formula to be used in MCTS. The author used Swiss tournament selection, reproduction and mutation to be applied in the 128 or 256 individuals. The author tested his approach using the game of Go. He demonstrated how his GP approach outperformed MCTS and RAVE when a relatively small number of play outs were used, but UCT outperformed his approach when more play outs were allowed.}

{Motivated by Cazenave's studies, a decade later, Bravi et al.~\cite{Bravi2017EvolvingGU} evolved the MCTS UCB1 formula tested in the General Video Game AI framework. They used a population of 100 GP trees and the initial population started with a seeded UCT individual and 99 random trees. In their studies, they considered three different scenarios: (a) only given access to the same information as UCB1, (b) given access to additional game-independent information, and (c) given access to game-specific information. The authors showed that, on average, Scenario (c) outperformed the rest of the scenarios including the UCB1.}

{Holmg{\aa}rd et al.~\cite{DBLP:journals/tciaig/HolmgardGLT19} used GP to evolve persona-specific evaluation formulae to be used in MCTS instead of the UCB1 in the deterministic game of MiniDungeons 2.  To evolve the 100 GP individuals over 100 generations, the authors used selection, crossover, mutation and elitism, as well as an island model. They reported that their evolved personas were able to play the game more efficiently compared to the UCB1 agents.}

{Other interesting works include those carried out by  Alhejali and Lucas~\cite{6633639}, where they used a  GP system to enhance MCTS during the roll out simulations tested in the game of Ms PacMan. The authors used multiple functions including comparison, conditional, logical operators. They used 100 or 500 individuals evolved, using selection and mutation operators, over 50 or 100 generations, where a single run took around 18 days to finish, for the latter case.

 Another interesting work is related to handling prohibitive branch factors in MCTS through EAs, as proposed by  Baier and Cowling~\cite{8490403} in the deterministic game of Hero Academy. Lucas et al.~\cite{Lucas2014} used an EA as a source of control parameters to bias the roll outs of MCTS. They showed how their proposed approach significantly outperforms the vanilla MCTS using the Mountain Car benchmark problem and a simplified version of Space Invaders.}

These works are inspiring, but the contributions of this work are novel and timely. 
\begin{enumerate}
\item Firstly, instead of using a single type of problem as normally done so far by the community, we use five functions of different nature and complexity, going from a unimodal function, covering multimodal functions, to deceptive functions.
\item Secondly, we use a state-of-the-art EA algorithm, inspired by semantics, to evolve a selection policy to be used instead of the MCTS UCT. 
\item Thirdly, instead of using a large population size and a large number of generations as done in all the previous works described before, we use a small population size evolved by a few generations to see if it is possible to successfully evolve a selection policy to be used in lieu of the MCTS UCT. 
\item Fourthly, we compare five different variants of the MCTS UCT \textit{vs.} our semantic-inspired EA and shed some light under what circumstances an evolved selection policy might have a positive effect on MCTS.
\end{enumerate}

The rest of this paper is organised as follows. Section~\ref{sec:background} provides some background in MCTS, EAs, semantics and in the functions used in this work. Section~\ref{sec:ai:controllers} discusses in detail the controllers used in this work. Section~\ref{sec:experimental} presents the experimental setup. Section~\ref{sec:results} discusses the results obtained by each of the controllers. Section~\ref{sec:conclusions} draws some conclusions.

\section{Background}
\label{sec:background}

\subsection{The Mechanics Behind MCTS}

MCTS relies on two key elements: (a) that the true value of an action can be approximated using simulations, and (b) that these values can be used to adjust the policy towards a best-first strategy. The algorithm builds a partial tree, guided by the results of previous exploration of that tree. Thus, the algorithm iteratively builds a tree until a condition is reached or satisfied (e.g., number of simulations, time given to  Monte Carlo simulations), then the search is halted and the best performing action is executed. In the tree, each node represents a state, and directed links to child nodes represents actions leading to subsequent states.
Like many AI techniques, MCTS has several variants. Perhaps, the most accepted steps involved in MCTS are those described in~\cite{6145622}  and are the following: (a) \textit{Selection}: a selection policy is recursively applied to descend through the built tree until an expandable (a node is classified as expandable if it represents a non-terminal state and also, if it has unvisited child nodes) node has been reached, (b) \textit{Expansion}: normally one child is added to expand the tree subject to available actions, (c) \textit{Simulation}: from the new added nodes, a simulation is run to get an outcome (e.g., reward value), and (d) \textit{Back-propagation}: the outcome obtained from the simulation step is back-propagated through the selected nodes to update their corresponding statistics.

Simulations in MCTS start from the root state (e.g., actual position) and are divided in two stages: when the state is added in the tree, a tree policy is used to select the actions (the selection step is a key element and it is discussed in detail later in this section). A default policy is used to roll out simulations to completion, otherwise. 

One element that contributed to enhance the efficiency in MCTS was the selection mechanism proposed in~\cite{10.1007/11871842_29}. The main idea of the proposed selection mechanism was to design a Monte Carlo search algorithm that had a small probability error if stopped prematurely and that converged to the optimal solution given enough time. That is, a selection mechanism that nicely balances exploration \textit{vs.} exploitation, explained in the following paragraphs.


\subsection{Upper Confidence Bounds for Trees}
As indicated previously, MCTS works by approximating `real' values of the actions that may be taken from the current state. This is achieved through building a search or decision tree. The success of MCTS depends heavily on how the tree is built and the selection process plays a fundamental role in this. One particular selection mechanism that has proven to be reliable is the UCB1 tree policy~\cite{10.1007/11871842_29}. Formally, UCB1 is defined as:


\begin{equation}
  UCT = \overline{X}_j + C \sqrt{\frac{2 \cdot ln \cdot n}{n_j}}
  \label{eq:uct}
  \end{equation}

\noindent where $n$ is the number of times the parent node has been visited, $n_j$ is the number of times child $j$ has been visited and $K > 0$ is a constant. In case of a tie for selecting a child node, a random selection is normally used~\cite{10.1007/11871842_29}.

Thus, this selection mechanism works due to its emphasis on balancing both exploitation (first part of Eq. 1) and exploration (second part of Eq. 1). 

\subsection{Evolutionary Algorithms}

Evolutionary Algorithms (EAs)~\cite{Back:1996:EAT:229867,EibenBook2003}, also known as Evolutionary Computation systems, refer to a set of stochastic optimisation bio-inspired algorithms that use evolutionary principles to build robust adaptive systems.  The key element to these algorithms is undoubtedly flexibility in allowing the practitioner to use elements from two or more different EAs techniques. Consequently, the boundaries between these approaches are no longer distinct allowing a more holistic EA framework to emerge, such as the one adopted in this research. EAs work with a population of $\mu$-\textit{encoded} (representation of the) potential solutions to a particular problem. Each potential solution, commonly known as an individual, represents a point in the search space, where the optimal solution lies. The population is evolved by means of genetic operators, over a number of generations, to produce better results to the problem.  Each individual is evaluated using a fitness function to determine how good or bad the individual is for the problem at hand. The fitness value assigned to each individual in the population probabilistically determines how successful the individual will be at propagating (part of) its code to future generations. 

The evolutionary process is carried out by using genetic operators.  Selection, crossover and mutation are the key operators used in most EAs. The selection operator is in charge of choosing one or more individuals from the population based on their fitness values. Multiple selection operators have been proposed. One of the most popular selection operators is tournament selection where the best individual is selected from a pool, normally of size = $[2-7]$, from the population. The stochastic crossover, also known as recombination, operator exchanges material normally from two selected individuals. This operator is in charge of exploiting the search space. The stochastic mutation operator makes random changes to the genes of the individual and is in charge of exploring the search space. The mutation operator is important to guarantee diversity in the population as well as recovering genetic material lost during evolution. This evolutionary  process is repeated until a stopping condition is reached such as until a maximum number of generations has been executed. The population, at this stage, contains the best evolved potential solutions to the problem and may also represent the global optimal solution.

The field has its origins in four landmark evolutionary methods: Genetic Algorithms~\cite{DBLP:books/aw/Goldberg89}, Evolution Strategies~\cite{Rechenberg10.1007/978-3-642-83814-9_6}, Evolutionary Programming~\cite{Fogel1966} and Genetic Programming~\cite{Koza:1992:GPP:138936}. In this work we briefly describe the two methods employed in this work: Evolution Strategies and Genetic Programming. The second author's work in Neuroevolution in Deep Neural Networks~\cite{9383028} provides a nice summary of all of these EAs. 

\subsubsection{Evolutionary Algorithm: Genetic Programming (GP)} This EA was  popularised by Koza~\cite{Koza:1992:GPP:138936}. GP is a form of automated programming where individuals are randomly created by using  functional and terminal sets required to solve a given problem. Multiple types of GP have been proposed in the literature with the typical tree-like structure being the predominant form of GP in EAs.

\subsubsection{Evolutionary Algorithm: Evolution Strategies (ES)} These EAs were introduced in the 1960s by Rechenberg~\cite{Rechenberg10.1007/978-3-642-83814-9_6}. ES are generally applied to real-valued representations of optimisation problems. In ES, mutation is the main operator whereas crossover is the secondary, optional, operator. Historically, there were two basic forms of ES, known as the ($\mu,\lambda$)-ES and the ($\mu+\lambda$)-ES. $\mu$ refers to the size of the parent population, whereas $\lambda$ refers to the number of offspring that are produced in the following generation before selection is applied. In the former ES, the offspring replace the parents whereas in the latter form of ES, selection is applied to both offspring and parents to form the population in the following generation. 

\subsection{Semantics}



For clarity purposes, we first briefly give some definitions on semantics, based on the first author's work~\cite{DBLP:conf/gecco/GalvanS19}, that will allow us to describe our approach later in Section~\ref{sec:ai:controllers}.

Let $p \in P$ be a program from a language $P$. When $p$ is applied to an input  $in \in I$, $p$ produces an output $p(in)$. 

\begin{mydef}
\indent  \textcolor{black}{The semantic mapping function $s: P \rightarrow S$  maps any program $g$ to its semantics $s(g)$.}
\label{def:general}
\end{mydef}

This means, $\textcolor{black}{s(g_1) = s(g_2) \Longleftrightarrow \forall in \in I : g_1(in) = g_2(in).}$ The semantics specified in Def.~\ref{def:general} has three properties. Firstly, every program has only and only one semantics. Secondly, two or more programs can have the same semantics. Thirdly, programs that produce different outputs have different semantics. Def.~\ref{def:general} is general as it does not specify how semantics is represented. This work is inspired by a popular version of semantics GP where the semantics of a program is defined as the vector of output values computed by this program for an input set (also known as fitness cases). The latter are not available in MCTS. We then extrapolate this idea to the fitness space. Thus, assuming we use a finite set of simulations, as normally adopted in MCTS, we can now define, without losing the generality, the semantics of a program in the simulations.

\begin{mydef}
  The semantics $s(p)$ of a program $p$ is the vector of values from each independent simulation $sim$,
\label{def:semantics}
  \end{mydef}

Thus, we have that the semantics of a program in MCTS is given by $s(p) = [p(sim_1), p(sim_2), \cdots, p(sim_l)]$,  where $l = |I|$ is the number of independent simulations.

Based on Def.~\ref{def:semantics}, we can define the \textit{Sampling Semantics Distance} (SSD) between two programs $(p,q)$. \textcolor{black}{That is, let $P$ = $\{p_1, p_2, ..., p_N \}$ and $Q$ = $\{q_1, q_2, \cdots, q_N \}$ be the sampling semantic of Program 1 ($p_1$) and Program 2 ($p_2$) on the same set of sample points, then the SSD between $p_1$ and $p_2$ is defined as} ${SSD(p,q) = (|p_1-q_1|+|p_2-q_2|+...+|p_N-q_N|)/N}$, where $S_p = \{p_1,...,p_N\}$ and $S_q = \{q_1,...,q_N\}$ are the SS of programs $p$ and $q$ based on simulations.

We are now in position to use the well-known semantic similarity (SSi) proposed by the first author and colleagues~\cite{DBLP:journals/gpem/UyHOML11}. This indicates whether the SSD between two programs lies between a lower bound $\alpha$ and upper bound $\beta$ or not. It determines if two programs are similar without being semantically identical. The SSi of two programs $p$ and $q$ on a domain is formally defined as

\begin{equation}
  \textcolor{black}{  SSi(p,q) = (\alpha < SSD(p,q) < \beta)}
  \label{eq:ssi}
\end{equation}

\noindent where $\alpha$ and $\beta$ are the lower and upper bounds for semantic sensitivity, respectively. In our work, we set these 5 and 10, respectively.

\subsection{Test Functions}

\begin{figure}[t] 
    \centering\includegraphics[width=0.9\columnwidth]{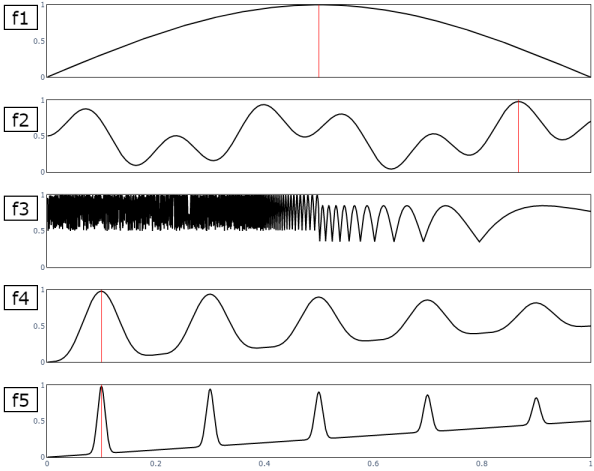}
    \caption{Functions used for the function optimisation problem.}
    \label{fig:functions}
\end{figure}

We use five different functions, each of different degree of difficulty, going from unimodal functions, including multimodal functions to highly deceptive functions. These functions are shown in Figure \ref{fig:functions}, and for simplicity, we constrain the domain and range of the functions to be in the interval $[0, 1]$.

The first function, shown in Figure~\ref{fig:functions}, f$_1$, depicts a unimodal function. The global optima is shown by a vertical red line. The function is defined by Eq.~\ref{eq:f1}, 

    \begin{equation}
      f_{1}(x) = \sin(\pi x)
      \label{eq:f1}
    \end{equation}

    The second function, shown in Figure~\ref{fig:functions}, f$_2$, is a multimodal function, with a global optima situated in the right-hand side of the figure as denoted by the vertical red line. The function is defined by Eq.~\ref{eq:f2},

    \begin{equation}
      f_{2}(x) =  0.5 \sin(13 x)  \sin(27 x) + 0.5
      \label{eq:f2}
    \end{equation}

   The third function, inspired by~\cite{james2017analysis}, is shown in Figure~\ref{fig:functions}, f$_3$. It shows a function with a high degree of ruggedness (left-hand side) containing multiple global optimum (not indicated as done with f$_1$ and f$_2$). In sharp contrast, the function also denotes smoothness in the right-hand side. The function is defined by Eq.~\ref{eq:f3},

    \begin{equation}
        f_{3}(x)=\begin{cases*}
            0.5 + 0.5 | \sin( \frac{1}{x^{5}})| & when $x<0.5$,\\
            \frac{7}{20} + 0.5 | \sin( \frac{1}{x^{5}})| & when $x\geq0.5$.
        \end{cases*} 
        \label{eq:f3}
    \end{equation}

    The fourth function used in this work is shown in Figure~\ref{fig:functions}, f$_4$. This is a deceptive function, where the global optimum is depicted by the red vertical line in the left-hand side of the figure. This function is defined by Eq.~\ref{eq:f4},
      
    \begin{equation}
        f_{4}(x) =  0.5 x + (-0.7 x+1)\sin(5 \pi  x)^{4}
        \label{eq:f4}
    \end{equation}

    The fifth and final function used in this work is shown in Figure~\ref{fig:functions}, f$_5$. This is also a deceptive function, slightly harder compared to f$_4$. This function is defined by Eq.~\ref{eq:f5},
   
    \begin{equation}
        f_{5}(x) =  0.5 x + (-0.7 x+1)\sin(5 \pi  x)^{80}
        \label{eq:f5}
    \end{equation}

\section{AI Controllers}
\label{sec:ai:controllers}

\subsection{Monte Carlo Tree Search}


The core idea of MCTS is based on its four functions: \textit{Selection}, \textit{Expansion}, \textit{Rollout} and \textit{Backpropagation}. The completion of these stages is known as a single simulation. When all simulations conclude, normally the node with the highest action value is chosen, which is the approach taken in this work. The MCTS is explained in detail in Section~\ref{sec:background}.



\subsection{Semantic-inspired Evolutionary Algorithm in MCTS}
\label{subsection:EA-p-MCTS}

We now turn our attention to the proposed AI controller based on EAs to evolve online the {mathematical} expression to be used during the selection phase of the MCTS. To this end, we use ($\mu$,$\lambda$)-ES (see Section~\ref{sec:background}).  We first use the UCT formula as parent  to later generate the offspring. We evolve a candidate solution in every turn that we need to make a decision. At each turn, a new evolved solution is built from scratch. Similar to~\cite{james2017analysis}, ``in the simulation phase, actions are executed uniformly randomly until a terminal state is encountered, at which point some reward is received. Let $f$ be the function and $c$ be the midpoint of the state reached by the rollout. At iteration $t$, a binary reward $r_t$, drawn from a Bernoulli distribution $r_t \sim Bern (f (c))$, is generated''. This is how the fitness is assigned to each evolved potential solution. The value of these  are used to update a \textit{copy} of the MCTS statistical tree, from the selected node to the root including the nodes given in a branch. We perform 30 simulations to compute the fitness of the evolved expression. The fitness of our evolved individual is the average of these  30 simulations. We pick the offspring based on semantics to act as parent.

A potential major limitation in using a small population size, as adopted in this work, could be the lack of diversity leading to poor performance. To prevent this, we use semantics as inspiration to promote diversity (see Section~\ref{sec:background}). We dubbed this method Semantic-inspired Evolution Algorithms in Monte Carlo Tree Search (\THIRDEA). 

{We first get the highest fitness, H$_f$, from the offspring . If there is more than one offspring with H$_f$, we compute the sampling semantic distance from each offspring with respect to Parent. We then proceed to compute the semantic similarity metric using thresholds. If there is more than one individual from the offspring population that falls within this threshold, defined by $\alpha$ and $\beta$, then the individual closest to the $\alpha$ value is picked. Otherwise, we select an individual randomly from the offspring population.}

Our proposed method aims to evolve  mathematical expressions that can replace UCT  with the goal to get better or competitive results compared to UCT. Thus, ES is called during the selection step in MCTS. Once a  node has been selected by our evolved expression, we proceed to compute the fitness of the evolved expression. We do so by performing roll outs as done in MCTS.

\section{Experimental Setup}
\label{sec:experimental}

\subsection{Function and Terminal Sets}
The terminal set is defined by  $T =  \{Q(s,a),N(s),N(s,a),K\}$, where $N(s)$ is the number of visits to the node from the MCTS search tree, $N(s,a)$ is the number of visits to a child node, $Q(s,a)$ is the child's node action-value and $K$ is the exploration-exploitation constant. When $K$ is chosen to be mutated, it can take a random value from the following set $r = \{0.5, 1, \sqrt2, 2, 3\}$. The function set is defined by $F = \{+,-,*,\div,\log,\sqrt { }\}$, where the division operator is protected against division by zero and will return 1 for any divisor less than 0.001. Similarly, the natural log and square root operators are protected by taking the absolute values of input values. The values used for all our controllers are shown in Table~\ref{tab:parameters}.

\subsection{Function Optimisation}

The test functions  used in this work, commonly known as  function optimisation as described in \cite{bubeck2010x,james2017analysis},  is a problem where each state $s$ represents a domain [$a_{s}$,$b_{s}$], starting at [0,1]. The available actions from state $s$ are $b$ evenly spaced partitions of the domain, sized $\frac{a_{s}-b_{s}}{b}$ each, where $b$ is the selected branching factor. The objective is to find the state where the global maxima of a given function $f$ lies. A state is said to be terminal if its domain is smaller than the threshold $t$, in other words $b_{s}-a_{s}<t$. The threshold is set as $t=10^{-5}$ and the branching factor is set as $b=2$.

The MCTS rollouts use a random uniform default policy. When a terminal state is reached, $f$ is evaluated at the state's central point $c_{s}=\frac{(a_{s}-b_{s})}{2}$. The reward $r_{s}$ can either be 1 or 0 and is sampled from a Bernoulli distribution $r_{s}\sim Bern(f(c_{s}))$. Thus, $0\leq f(x)\leq 1 | x \in [0,1] $ is ensured for every function $f$.


\begin{table}[tb]
\centering
\caption{MCTS agents parameters.}
\resizebox{0.90\columnwidth}{!}{ 
\small\begin{tabular}{|l|r|} \hline 
\emph{Parameter} & \emph{Value} \\ \hline \hline

\multicolumn{2}{|c|}{All MCTS agents}\\ \hline
C & {$\frac{1}{2}, 1, \sqrt2, 2, 3$} \\ \hline
Rollout playouts & 1 \\ \hline
\MCTSIts & {5000} \\ \hline

\multicolumn{2}{|c|}{\SIEAMCTS}\\ \hline
($\mu$,$\lambda$)-ES &  $\mu=1$, $\lambda=4$  \\ \hline
Generations & 20 \\ \hline
Type of Mutation & Subtree (90\% internal node, 10\% leaf) \\ \hline
Initialisation Method & Initial formula: UCB1 \\ \hline
Maximum depth & 8 \\ \hline
\MCTSIts & 2,400 EA evaluations + 2,600 iterations \\ \hline

\end{tabular}
}
\label{tab:parameters}
\end{table}

\section{Discussion of Results}
\label{sec:results}

\begin{figure}[tb] 
    \centering\includegraphics[width=0.9\columnwidth]{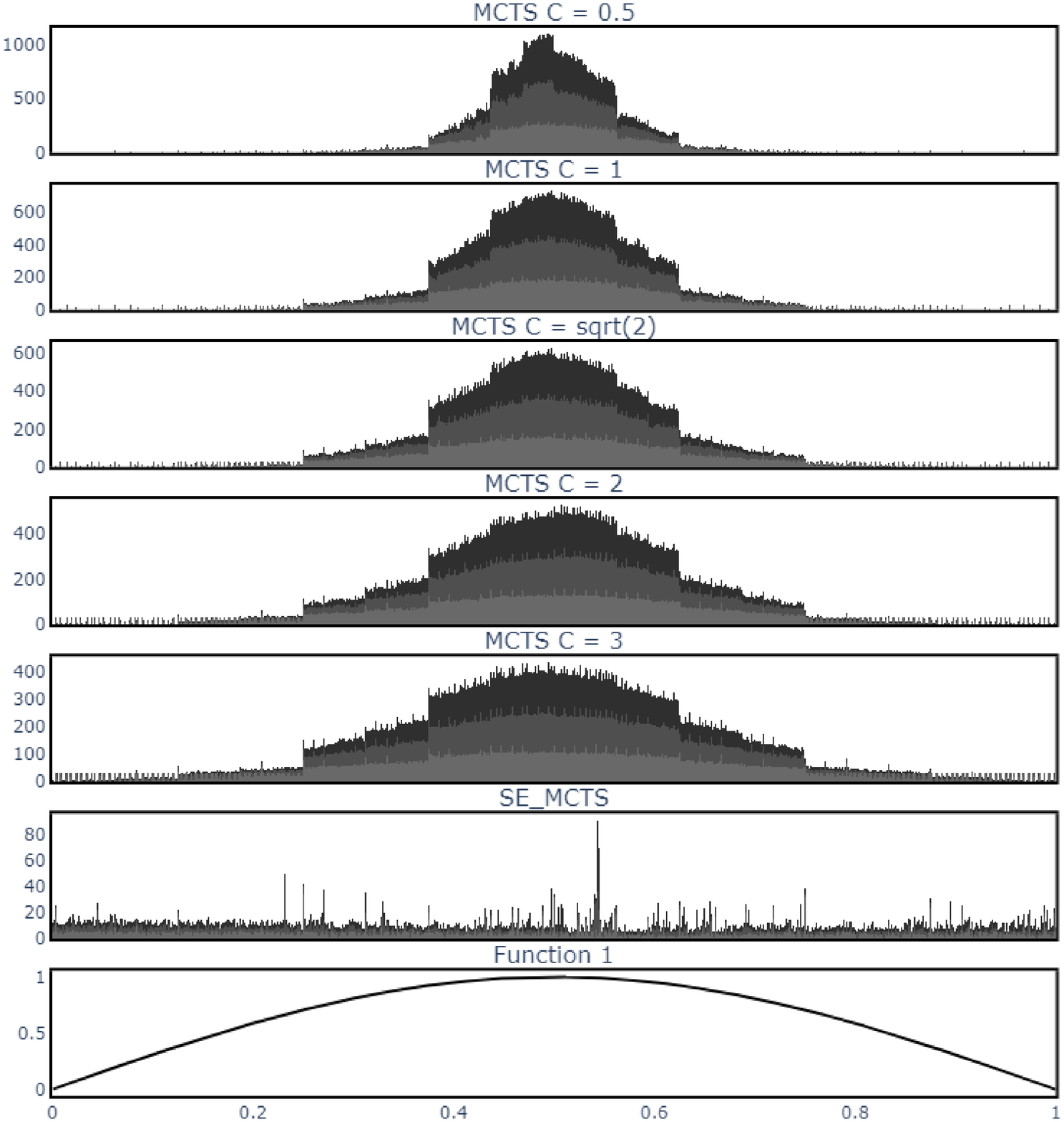}
    \caption{Histogram of the location of the nodes for Function 1.}
    \label{fig:f1}
\end{figure}

\begin{figure}[tb] 
    \centering\includegraphics[width=0.9\columnwidth]{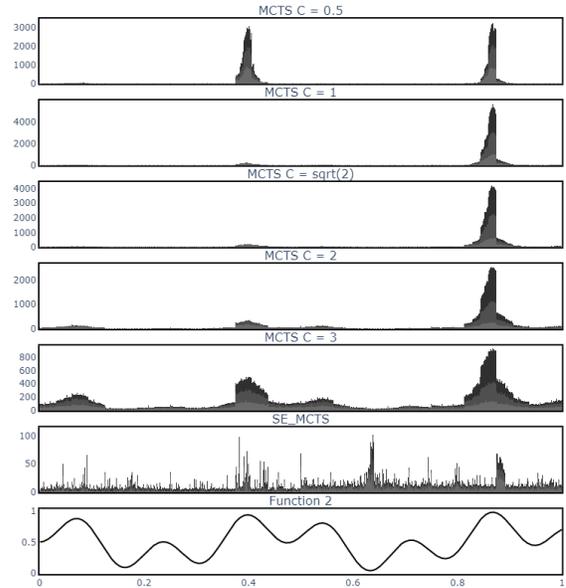}
    \caption{Histogram of the location of the nodes for Function 2.}
    \label{fig:f2}
\end{figure}

We are interested in knowing the effects of MCTS using different values of the UCT C constant and compared them with evolved functions that are used in the selection policy in lieu of the UCT formula. To do so, we keep track of how the search is carried out during the 5,000 iterations used by the MCTS and 2,600 iterations used by the EA (plus 2,400 evaluations). Thus, we will naturally observe a less number of nodes expanded in particular regions of a given formula when using the EA approach.

\begin{figure}[tb] 
    \centering\includegraphics[width=0.9\columnwidth]{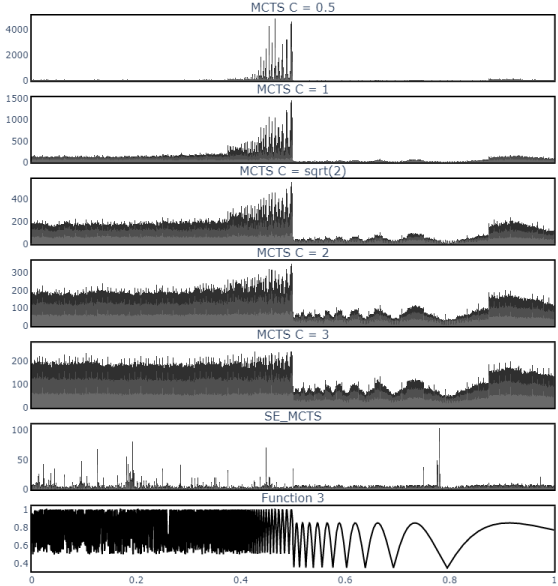}
    \caption{Histogram of the location of the nodes for Function 3.}
    \label{fig:f3}
\end{figure}

\begin{figure}[t] 
    \centering\includegraphics[width=0.9\columnwidth]{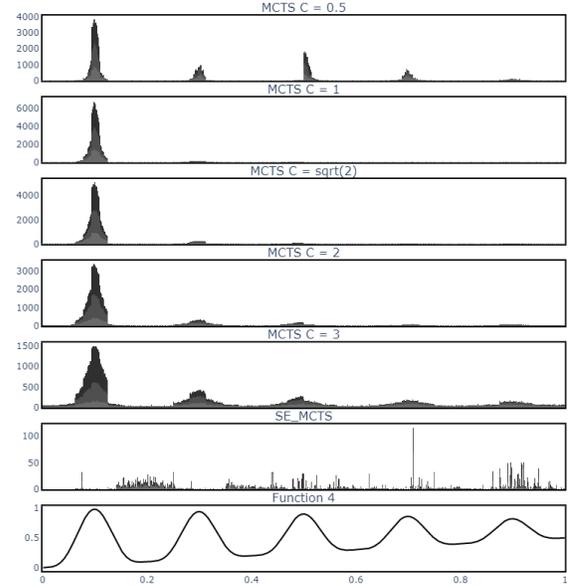}
    \caption{Histogram of the location of the nodes for Function 4.}
    \label{fig:f4}
\end{figure}

\begin{figure}[t] 
    \centering\includegraphics[width=0.9\columnwidth]{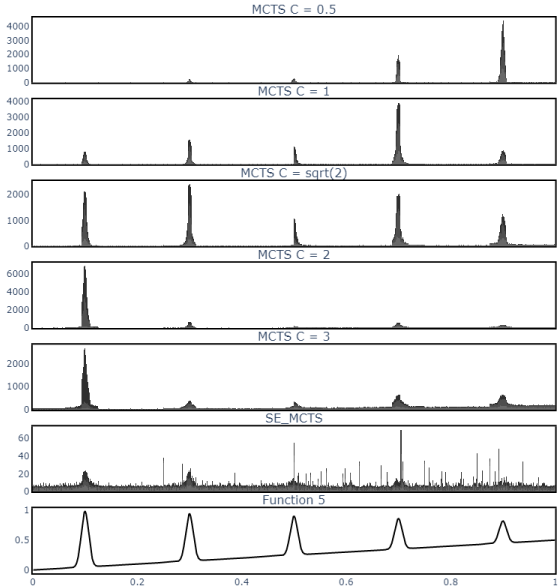}
    \caption{Histogram of the location of the nodes for Function 5.}
    \label{fig:f5}
\end{figure}

We divided these iterations equally by three. The first part is plotted with a light grey colour in Figures~\ref{fig:f1}--\ref{fig:f5}, the second part of this number of iterations is represented by a darker grey colour. Finally, the last part of is represented in black. In these figures, the $x$-axis correspond to the domain of the function and the $y$-axis corresponds to the number of nodes expanded in a particular region. The results are averaged from 30 independent runs for each AI controller. The histograms are generated by sorting the nodes of the tree into bins, according to the location of the center of their states. The goal is to illustrate how and when the nodes are expanded in different regions of the domain. 

Let us start our analysis by analysing our first function (see Section~\ref{sec:background}), re-plotted at the end of Figure~\ref{fig:f1} for convenience. This is an unimodal function and the easiest to be solved. The first five plots, from top to bottom, correspond to MCTS using C=\{0.5, 1, $\sqrt2$, 2, 3\}. As can be seen, as the C value increases, the MCTS explores larger regions. For instance, see the fifth plot, where C = 3, compared to the first plot, where C = 0.5. In the former case, there are close to 400 nodes around the global optima, compared to almost 1,000 nodes when using the latter. Let us now turn our attention to our EA method. We can see in plot sixth, from top to bottom, that this method has more variations compared to MCTS, regardless of the UCT C value used. This is expected given that at each node selection, we evolve an alternative UCT formula. Despite this, it is interesting to note that our EA method is able to expand a good number of nodes close to the global optima (center of the Function 1 as seen at the bottom of Figure~\ref{fig:f1}).

Let us continue with our second function (see Section~\ref{sec:background}), seen at the bottom of Figure~\ref{fig:f2}. This multimodal function has also a global optima located in the right-hand side of the figure. When we focus our attention on how the MCTS behaves using this function (see first five plots from top to bottom in Figure~\ref{fig:f2}), we can observe that as C is higher, the MCTS tends to expand nodes in the global optima as well as in local optima. This is depicted in the fifth plot, where there are three peaks. This situation changes as the UCT C value decreases, except when UCT C = 0.5, where two peaks are formed, one of them being where the global optima lies. Another point worth noting is the fact that the MCTS UCT C = \{1, $\sqrt2$\} exhibit a similar behaviour: both expand nodes around the global optima region as well as expanding a small number of nodes in a local optima region. When we turn our attention to the EA, we can see again a large variation in exploring/exploiting the search space. It has around seven peaks, one of them formed around the global optima region. 

We now focus our attention on the third function used in this work (see Section~\ref{sec:background}) and depicted at the bottom of Figure~\ref{fig:f3}. This shows a high degree of  rugedeness, left-hand side of the last plot, where multiple global optimum exist. In sharp contrast, this function also denotes smoothness in the right-hand side. As can be seen in the first five plots, from top to bottom, of Figure~\ref{fig:f3}, MCTS with a high UCT C value tends to explore more and so, the nodes tend to expand in `incorrect’ areas of the search space, as seen when C = \{2, 3\}. This is also visible when C = \{$\sqrt2$\}. The situation changes dramatically as UCT C decreases, for instance, see the first plot, where UCT C = 0.5. In this case there is a good number of nodes expanded in a narrow region where the global optimum lie. When we turn our attention to how the EA behaves, we can see an interesting trend: there is a good number of peaks, albeit small, nicely spread in the correct region where multiple global optimum are. This is in contrast to what is observed when using MCTS UCT C = 0.5, where some peaks are formed in just a fraction of the `correct’ part of the search space.

Let us focus our attention on the fourth function, defined in Section~\ref{sec:background} and plotted at the bottom of Figure~\ref{fig:f4}. This is a deceptive function, with a global optimum located in the left-hand side. Most of the MCTS perform well on this type of function. Although it is fair to say that MCTS UCT C = 1 outperforms the rest of the UCT C values as it has the highest number of nodes expanded where the global optimum lies. When we proceed to analyse the behaviour of the EA approach, we can see that this exhibits some interesting behaviours. For example, it is able to spend some time in the region where the global optimum lies, although this is only a single line with a few number of nodes (around 35). The EA also spends some time in local optima as shown by the peaks formed as a result of the number of expanded nodes. 

We finally proceed to focus our attention on the fifth function, defined in Section~\ref{sec:background}, shown at the bottom of Figure~\ref{fig:f5}. This is a deceptive function, harder compared to the fourth function, also deceptive, analysed previously. We can see that now the MCTS UCT C = 2 performs better compared to the rest of the UCT C values used, since it is able to spend more time in the global optimum located in the left-hand side of the function (bottom of Figure~\ref{fig:f5}). This is in sharp contrast to what we observed in Function 4, as aforementioned, MCTS UCT C = 1 performed better in this type of function. When we proceed to analyse the results of the EA, sixth plot from top to bottom of Figure~\ref{fig:f5}, we can see that now the algorithm manages to spend more time in the correct area compared to the behaviour shown in Function 4 (a deceptive function too with smooth areas). Although it is fair to say that the EA also spends time in local optima, as shown by the peaks formed in other parts of the function. 

From this analysis, we can see that MCTS UCT behaves very well in f$_1$ and f$_2$. For the rest of the functions, it also performs well, although a correct UCT C value is needed as discussed before. For example, we have argued that for a deceptive function with some degree of smoothness (f$_4$), the MCTS UCT = 1 behaves very well, but for a harder deceptive function (f$_5$), the MCTS UCT C = 2 is preferred. Perhaps this is one of the reasons why some researchers have carried out some research in evolving a selection policy that can be used in lieu of the MCTS UCT. When we observe the behaviour of the EA on the evolution of an alternative formula, we can see that this attains a poorer performance with MCTS UCT. However, it should be noted that in some cases, the use of the EA in MCTS can have a positive effect, such as in Functions 3 (high degree of ruggedness containing multiple global optima) and 5 (highly deceptive function).

\section{Conclusions}
\label{sec:conclusions}

Monte Carlo Tree Search (MCTS) is a sampling best-first method to search for optimal decisions. The success of MCTS depends heavily on how the MCTS statistical tree is built and the selection policy plays a fundamental role in this. A MCTS selection policy that works particularly well is the Upper Confidence Bounds for Trees, referred to as UCT. However, some tuning is necessary for this to work well. Moreover, some sophisticated bounds have been proposed by the community to be used in lieu of the MCTS UCT. 

As a result of this, some works e.g., ~\cite{Cazenave2007EvolvingMT,Bravi2017EvolvingGU,DBLP:journals/tciaig/HolmgardGLT19} have proposed evolving the MCTS UCT in hope to get a better performing MCTS. Although all these works are inspiring, the entirety of these have focused their attention on a particular problem rather than a set of problems with certain features. This has limited generalising their findings and knowing under what circumstances the evolution of the MCTS UCT might be beneficial.

In this work, we have shown how the evolution of the MCTS UCT might be beneficial in multimodal scenarios as well as in deceptive ones. In contrast to this, the MCTS UCT performs incredibly well in unimodal scenarios and is competitive in the rest of the scenarios. Thus, we argue that it is important to know the features of the problem at hand when attempting to evolve the MCTS UCT to be used in lieu of the UCT commonly adopted in Monte Carlo Tree Search.

\section*{Acknowledgments}
This work has emanated from research conducted with the financial support of Science Foundation Ireland (SFI) under Grant Number SFI 18/CRT/6049.

\bibliographystyle{abbrv}
\bibliography{evolvingUCT_arXiv}

\end{document}